\documentclass[letterpaper]{article} 
\usepackage{aaai25}  
\usepackage{times}  
\usepackage{helvet}  
\usepackage{courier}  
\usepackage[hyphens]{url}  
\usepackage{graphicx} 
\usepackage{natbib}  
\usepackage{caption} 
\usepackage{algorithm}
\usepackage{algorithmic}
\usepackage{amsfonts}
\usepackage{amsmath}
\usepackage{threeparttable}
\usepackage{enumitem}
\usepackage{booktabs}
\usepackage{newfloat}
\usepackage{listings}
\DeclareCaptionStyle{ruled}{labelfont=normalfont,labelsep=colon,strut=off} 
\floatstyle{ruled}
\newfloat{listing}{tb}{lst}{}
\floatname{listing}{Listing}
\title{DutyTTE: Deciphering Uncertainty in Origin-Destination Travel Time Estimation}
\author{
  Xiaowei Mao\textsuperscript{\rm 1, 2},
     ~Yan Lin\textsuperscript{\rm 3},
     ~Shengnan Guo\textsuperscript{\rm 1, 2},
     ~Yubin Chen\textsuperscript{\rm 1, 2},
     ~Xingyu Xian\textsuperscript{\rm 1, 2},\\
     ~Haomin Wen\textsuperscript{\rm 4},
     ~Qisen Xu\textsuperscript{\rm 1, 2},
     ~Youfang Lin\textsuperscript{\rm 1, 2},
     ~Huaiyu Wan\textsuperscript{\rm 1, 2}
}
 \affiliations {
     \textsuperscript{\rm 1}School of Computer and Information Technology, Beijing Jiaotong University, Beijing, China\\
     \textsuperscript{\rm 2}Beijing Key Laboratory of Traffic Data Mining and Embodied Intelligence
, Beijing, China\\
    \textsuperscript{\rm 3}Aalborg University, Aalborg, Denmark\\
        \textsuperscript{\rm 4}Carnegie Mellon University, Pittsburgh, USA\\
     
     \{maoxiaowei, guoshn, 20231244, xingyuxian, 23218133, yflin, hywan\}@bjtu.edu.cn;\\
     lyan@cs.aau.dk;\\
     wenhaomin.whm@gmail.com

}

\begin{document}
\maketitle

\begin{abstract}
Uncertainty quantification in travel time estimation (TTE) aims to estimate the confidence interval for travel time, given the origin (O), destination (D), and departure time (T). Accurately quantifying this uncertainty requires generating the most likely path and assessing travel time uncertainty along the path. This involves two main challenges: 1) Predicting a path that aligns with the ground truth, and 2) modeling the impact of travel time in each segment on overall uncertainty under varying conditions.
We propose DutyTTE to address these challenges. For the first challenge, we introduce a deep reinforcement learning method to improve alignment between the predicted path and the ground truth, providing more accurate travel time information from road segments to improve TTE. For the second challenge, we propose a mixture of experts guided uncertainty quantification mechanism to better capture travel time uncertainty for each segment under varying contexts. Extensive experiments on two real-world datasets demonstrate the superiority of our proposed method.
\end{abstract}

%
\begin{links}
    \link{Code}{https://github.com/maoxiaowei97/DutyTTE}
\end{links}

\section{Introduction}
Origin-destination (OD) travel time estimation (TTE) refers to estimating the travel time $\Delta t$ from the origin (O) to the destination (D) starting from the departure time (T). In many real-world applications, providing just an average estimate of travel time is inadequate. A more reliable and informative approach includes quantifying travel time uncertainty. Specifically, we aim to estimate the confidence interval for travel time with a specified confidence level to quantify uncertainty, which is beneficial in many scenarios. For example, ride-hailing services can benefit from providing customers with lower and upper confidence bounds of travel time, allowing them to better plan their schedules~\cite{liu2023uncertainty}. Moreover, understanding travel time uncertainty can help ride-hailing and logistics platforms improve decision-making effectiveness, such as in order dispatching and vehicle routing~\cite{xu2018large, CompactETA}. Existing studies for OD TTE~\cite{DeepOD, DOT-SIGMOD23, wang2023multi} primarily focus on point estimates of travel time and have not adequately addressed uncertainty quantification.

Quantifying origin-destination travel time uncertainty involves two key tasks. First, a path between OD that matches the ground truth needs to be predicted, as the accuracy of the path is closely related to the accuracy of the quantified uncertainty in TTE. Second, the travel time uncertainty in each segment along the predicted path needs to be quantified so that the travel time for the whole trip can be aggregated~\cite{DOT-SIGMOD23, wang2023multi}. Effectively accomplishing these tasks faces several challenges.

\begin{figure}[!t]
    \centering
    \includegraphics[width= 1\linewidth]{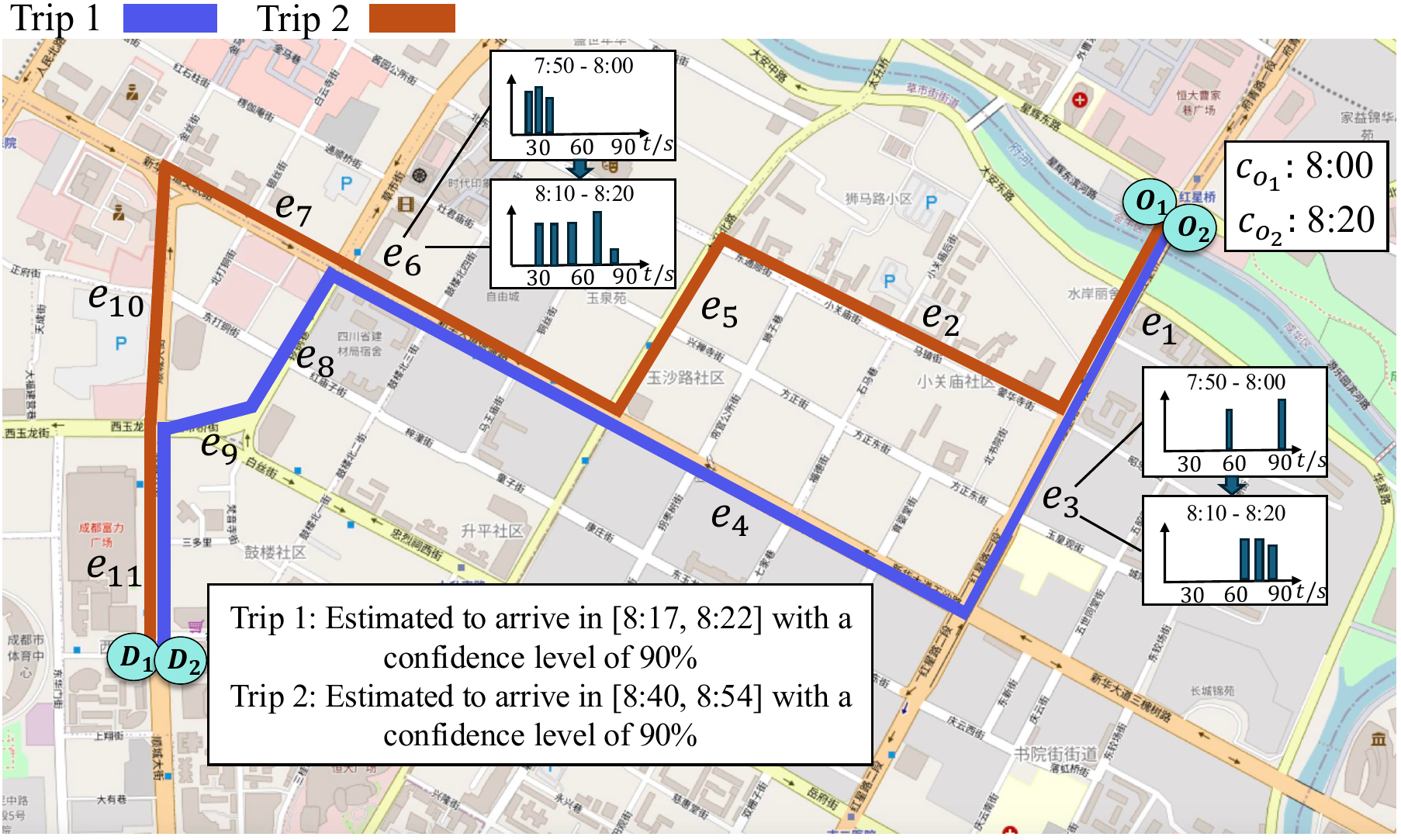}

        \caption{Motivation for DutyTTE: Each segment has varied travel time distributions, calculated using data from a period prior to the departure time.}
    \label{fig:intro}
   \vspace{-15pt}    
\end{figure}
\textbf{First, predicting a path that aligns with the ground truth is challenging.} Given an OD input, errors at multiple steps during path prediction can lead to significant divergence between the predicted and actual paths, especially for long paths with distant OD pairs. Such errors can accumulate, making the predicted path diverge from the actual path, which greatly hurts the accuracy of the quantified uncertainty in TTE. Traditional methods for path prediction typically involve modeling the weights of road segments, which are then integrated into search-based algorithms~\cite{jain2021neuromlr, tian2023effective} to generate a path step by step. More recent approaches~\cite{GDPICLR24, mao2023drl4route} use deep neural networks to learn transition patterns among roads and generate paths autoregressively. These methods face challenges in predicting paths that closely match the ground truth due to limited solutions for mitigating error accumulation during path prediction.

\textbf{Second, effectively modeling the impact of travel time in each road segment on overall travel time uncertainty under varying conditions is challenging}. Travel time in a segment varies across different trips and affects overall travel time uncertainty differently. As illustrated in Figure~\ref{fig:intro}, the travel time of \(e_6\) in trip 1, which starts at 8:00, is relatively certain since the statistical times during 7:50–8:00 are concentrated around 30 seconds. However, during 8:10–8:20, travel times of \(e_6\) become more varied, increasing the uncertainty for trip 2 which starts at 8:20. The varying travel time uncertainty in each segment, influenced by complex factors like traffic conditions and departure times, makes it difficult to accurately quantify overall travel time uncertainty.

To address these challenges, we propose DutyTTE for \underline{D}eciphering \underline{U}ncertain\underline{ty} in origin-destination \underline{T}ravel \underline{T}ime \underline{E}stimation. First, to reduce error accumulation in path prediction, we optimize path prediction by enhancing the overall alignment between the predicted path and the ground truth, rather than just maximizing the likelihood of each segment independently. We introduce a Deep Reinforcement Learning (DRL) approach to optimize objectives that measure the overall similarity between the predicted path and the ground truth. Second, we propose a Mixture of Experts guided Uncertainty Quantification mechanism (MoEUQ) to model the impact of travel time in each road segment on overall travel time uncertainty. This mechanism adaptively selects the most suitable experts to process each road segment, handling its travel time uncertainty and its impact in different contexts. This allows for better discrimination and capture of each road segment's impact on overall travel time uncertainty. 

Our contributions are summarized as follows:
\begin{itemize}
    \item We propose an OD travel time uncertainty quantification method that accurately predicts paths and provides reliable confidence intervals.
    \item We introduce a DRL method to enhance the alignment between the predicted path and the ground truth, facilitating accurate travel time uncertainty quantification.
    \item We present a mixture of experts guided uncertainty quantification mechanism to model the impact of travel time in each road segment on overall travel time uncertainty. 
    \item Extensive experiments on two real-world datasets show that our proposed method outperforms other solutions.
\end{itemize}

\section{Related Work}
\textbf{Travel Time Estimation}. 
TTE solutions are classified into two types: for paths and for OD input.

\par Path-based TTE methods like Wide-Deep-Recurrent (WDR)~\cite{wang2018learning}, DeepTTE~\cite{wang2018will}, and ProbTTE~\cite{liu2023uncertainty} use deep learning to model spatio-temporal correlations in traveled segments. DeepGTT~\cite{li2019learning} uses a variational model for TTE. GNNs and attention mechanism are used to capture spatial-temporal dependencies in road networks for TTE~\cite{fang2020constgat, hong2020heteta, derrow2021eta, CompactETA, jin2022stgnn, chen2022interpreting, yuan2022route, james2021citywide, ye2022cateta, liu2024lighttr, mao2023gmdnet}. ProbTTE~\cite{xu2024link} models link travel time with a Gaussian hierarchical model.

\par OD-based TTE methods, like TEMP~\cite{wang2019simple}, ST-NN~\cite{jindal2017unified}, MURAT~\cite{li2018multi}, DeepOD~\cite{yuan2020effective} model the correlations between ODT and travel times. DOT~\cite{DOT-SIGMOD23} develops a diffusion model for TTE using pixelated trajectories. MWSL-TTE~\cite{wang2023multi} proposes to search a path first, and sums the estimated travel times of segments. However, these methods face limitations in modeling the complex correlations between OD pairs and travel time under varying traffic conditions, and they lack the capability to effectively model travel time uncertainty in road segments.

\par \noindent \textbf{Uncertainty Quantification in Deep Learning}. Methods for quantifying uncertainty in deep learning can be broadly classified into Bayesian and frequentist approaches~\cite{wu2021quantifying}. Bayesian methods~\cite{pmlr-v139-izmailov21a,kendall2017uncertainties} quantify uncertainty by assigning probabilistic distributions to model parameters, with variations in these parameters representing uncertainty. Non-Bayesian methods include ensemble-based approaches~\cite{lakshminarayanan2017simple,liu2020probabilistic}, which model uncertainty by aggregating predictions from diverse models, and Brownian Motion-based techniques~\cite{kong2020sde}, which simulate stochastic dynamics in the learning process to capture uncertainty. Techniques like MIS-Regression~\cite{gneiting2007strictly} and Quantile Regression~\cite{gasthaus2019probabilistic} estimate uncertainty intervals by modifying output layers without relying on likelihood functions. These uncertainty quantification methods can be integrated with TTE approaches to estimate travel time uncertainty~\cite{zhu2022cross}. However, such combinations fail to accurately predict paths and account for the varying uncertainty of travel times across segments under complex conditions.

\section{Preliminaries}
In this section, we provide the relevant definitions and formalize the problem of uncertainty quantification in origin-destination travel time estimation.

\par \noindent {\textbf{Definition\hspace{3px}1.}} \textit{\textbf{Road Network.}} 
A road network is defined as the aggregate of all road segments in a studied area or a city. It is modeled as a directed graph $\mathcal G=(\mathcal V, \mathcal E)$, where $\mathcal V$ is a set of nodes $v_i$ representing road intersections or segment ends, and $\mathcal E$ is a set of edges $e_i$ representing road segments. Each node in the network has a unique index.

\par \noindent {\textbf{Definition\hspace{3px}2.}} \textit{\textbf{Trips and Paths.}} A trajectory $\mathcal T$ is a sequence of GPS points with timestamps: $\mathcal T= \langle (g_1, c_1),\dots, (g_{|\mathcal T|}, c_{|\mathcal T|}) \rangle$, where $g_i=(\mathrm{lng}_i,\mathrm{lat}_i), i=1,\dots,|\mathcal T|$ denotes $i$-th GPS point, and $|\mathcal T|$ denotes the total number of GPS points in the trajectory. After map-matching using OpenStreetMap, the trajectory of a trip is $\boldsymbol{x}_{\mathcal T} = \{(v_1, c_1), \cdots, (v_k, c_k)\}$, with the time index $c$ monotonically increasing. The total travel time of the trip is $y = c_k - c_1$. Note that multiple GPS data points can be located on the same road segment. A path is defined as a sequence of nodes $\boldsymbol{x} = (v_1, ..., v_{|\boldsymbol{x}|})$, where each pair of nodes is adjacent, i.e., $\forall i = 0, 1, \cdots, |\boldsymbol{x}|$, $(v_i, v_{i+1}) \in \mathcal E$.

\par \noindent {\textbf{Problem Statement.}} We aim to learn a mapping function $f_{\theta}$ to estimate the upper and lower confidence bounds for travel time based on an ODT input, denoted as $\hat{l}$, and $\hat{u}$, which quantifies the uncertainty of the travel time. Such that $[\hat{l}, \hat{u}]$ is a confidence interval that covers the actual travel time $\tau$ with a confidence level of $1-\rho$. We denote the ODT input as $\boldsymbol{q}=(g_o, g_d, c_o)$, where \(g_o\), \(g_d\), and \(c_o\) are the GPS points of the origin, destination, and departure time, respectively.

\section{Methodology}
\begin{figure*}[hbtp]
		\centering
		\includegraphics[width=1\linewidth]{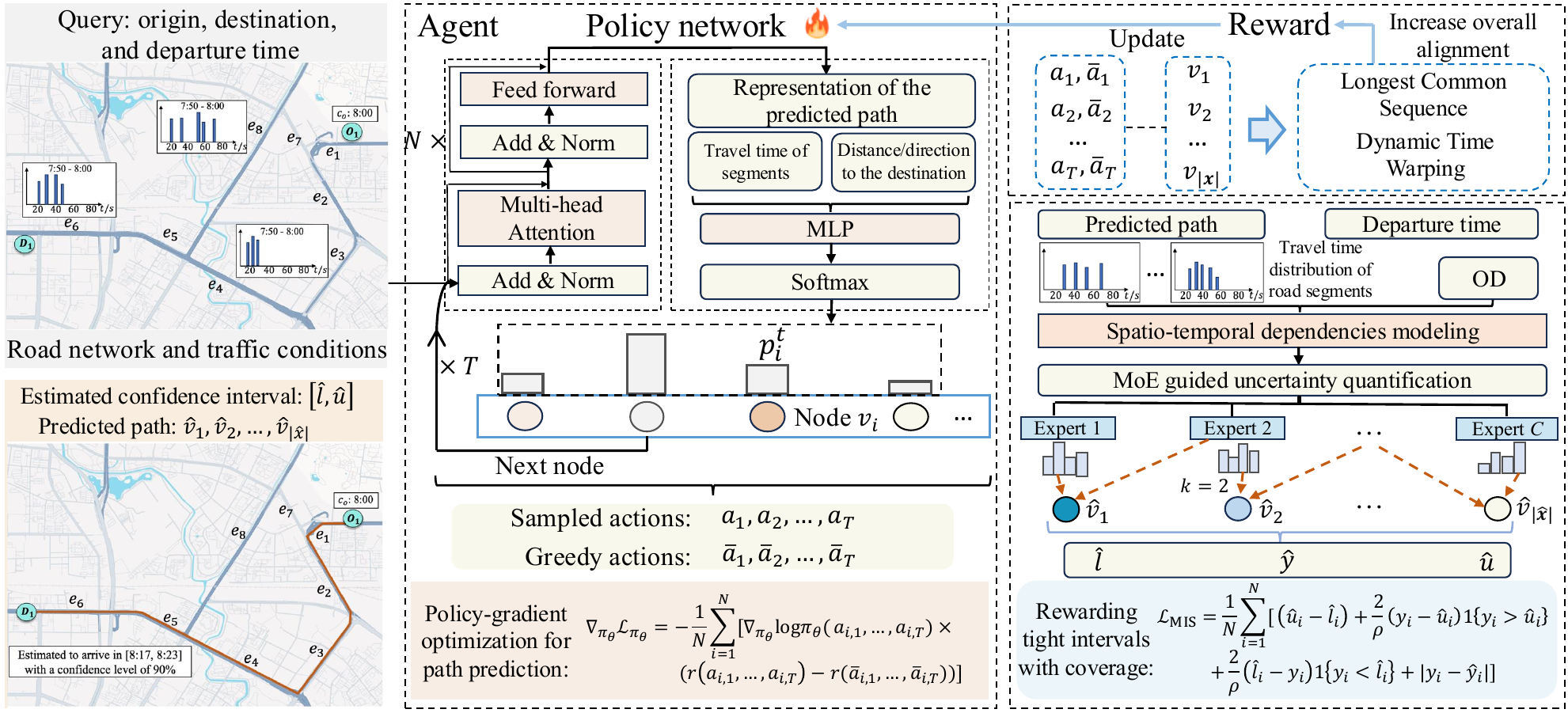}

   \caption{DutyTTE first learns to predict paths from an ODT input by optimizing objectives that measure the overall alignment. It then uses travel time information from segments in the predicted path to estimate travel time confidence intervals.}
  
		 \vspace{-15pt} 
		\label{fig_model}
\end{figure*}
Figure~\ref{fig_model} illustrates the proposed method. This section discusses path prediction from a DRL perspective, policy-gradient optimization to align predicted paths with ground truth, and MoE-guided uncertainty quantification for travel time.

\subsection{Path prediction from the DRL perspective}\label{DRL-PATH}
To mitigate error accumulation in path prediction, we aim to optimize overall alignment objectives between the predicted path $\hat{\boldsymbol{x}}$ and the ground truth ${\boldsymbol{x}}$. However, overall alignment measures like Longest Common Subsequence (LCS) and Dynamic Time Warping (DTW) are non-differentiable when used as loss functions due to the arg max operation involved in obtaining the predicted path. To address this, we introduce a deep reinforcement learning (DRL) method to enhance path prediction by improving overall alignment.

\par Path prediction is modeled as a finite-horizon discounted Markov Decision Process (MDP), where an agent, over $T$ discrete time steps, engages with the environment by making actions for path generation. An MDP is defined as $M = (\emph{S}, \emph{A}, \emph{P}, \emph{R}, \boldsymbol{s}_0, \gamma)$, with $\emph{S}$ being the set of states and $A$ being the set of possible actions. Next, $\emph{P}: \emph{S} \times \emph{A} \times \emph{S} \rightarrow {\mathbb R}_{+}$ is a transition probability function and $\emph{R}: \emph{S} \times \emph{A} \rightarrow {\mathbb R}$ is a reward function. Finally, $\boldsymbol{s}_0: S \rightarrow \mathbb{R}_{+}$ is the initial state distribution and $\gamma \in [0, 1]$ is a discount factor.
\par Given a state $\boldsymbol{s}_t$ at time $t$, the agent uses the policy $\pi_{\theta}$ to make sequential decisions regarding the next node to generate. The agent then receives a reward $r_t$. Next, the reward is used to update the policy network. The objective is to learn the optimal parameters $\theta^{*}$ for the policy network to maximize the expected cumulative reward:   \(\theta^{*} = {\arg\max}_{\theta}{\mathbb{E}_{\pi_{\theta}}\left[\mathop\sum\limits_{t=1}^{T}{\gamma^t}r_t\right]} \), where $\gamma$ controls the tradeoffs between the importance of immediate and future rewards. Next, we describe agent, state, action, reward, and state transition in detail.

\par \textbf{Agent.} The agent takes actions according to the policy based on the states and receives rewards to learn an optimal policy that maximizes the cumulative reward. The actions are determined by a policy network. The policy network is responsible for learning the mapping function $f_{\theta}$ from an ODT to a path. In our formulation, the policy network is mainly composed of a transformer~\cite{vaswani2017attention} and a prediction head. The transformer generates the representation of the predicted path. Next, the prediction head takes the representation of the predicted path, traffic conditions, distance and direction to the destination as input, then it outputs the distribution of generating the next node.

\par \textbf{State.} The state at time step $t$, denoted as $s_t \in \mathcal{S}$, encodes context information necessary to derive a policy at each time step and is given by $\boldsymbol{s}_t = (\boldsymbol{h}_t, {\hat{\boldsymbol{x}}}_{1: t-1}, {w}_{t}, dis_{t}, dir_{t})$. Here, \(\boldsymbol{h}_t\) denotes the representation of the predicted path, \({\hat{\mathbf{x}}}_{1: t-1}\) is the previously predicted path, \(w_t\) denotes traffic conditions, \(dis_{t}\) is the distance from the current node to the destination, and \(dir_{t}\) is the direction from the current node to the destination at the $t$-th time step.

\textbf{Action.} An action $a_t \in \mathcal{A}_t$ represents choosing the next node to generate in the path. If there are \(T\) nodes in a path, the sequence $(a_1, a_2 \cdots, a_{T}) \in \mathcal{A}_1 \times \mathcal{A}_2 \times \cdots \times \mathcal{A}_{T}$ represents a predicted path.

\par \textbf{State transition probability.} The transition probability function $P(\boldsymbol{s}_{t+1}|\boldsymbol{s}_t, a_t): \emph{S} \times \emph{A} \times \emph{S} \rightarrow {\mathbb R}_{+}$ captures the probability of transitioning from $\boldsymbol{s}_t$ to state $\boldsymbol{s}_{t+1}$ when action $a_t$ is taken. In our problem, the environment is deterministic, meaning there is no uncertainty in the state transition. Therefore, state $\boldsymbol{s}_{t+1}$ is deterministic.

\par \textbf{Reward.} The reward, aimed at improving the alignment between the predicted path and the ground truth, is defined and detailed in the following section.

\subsection{Policy-gradient optimization for path prediction}
\par Maximize the expected cumulative reward can be converted to a loss function defined as the negative expected reward of the predicted path: \( {\mathcal L_{\theta}}=- \mathbb{E}_{\pi_{\theta}}[r(a_1,\cdots,a_T)]\), where \(\{a_1, \cdots, a_T\} \sim \pi_{\theta}\) are sampled from the policy \(\pi_{\theta}\) and \(r(a_1, \cdots, a_T)\) represents the reward. This reward is defined by overall alignment objectives, utilizing metrics such as DTW and LCS. DTW calculates the cost to align nodes of the predicted path with the ground truth in a way that minimizes the total distance. While LCS identifies the longest subsequence common to both the predicted path and the ground truth. Generating paths with lower DTW and higher LCS values indicates greater similarity to the ground truth. Formally, the reward function calculated using the predicted path $\hat{\boldsymbol{x}}$ and the ground truth $\boldsymbol{x}$ is formulated as follows:
\begin{equation}
\begin{aligned}
r(a_1,\cdots,a_T) = (\omega * \rm{LCS}(\hat{\boldsymbol{x}}, \boldsymbol{x}) - \beta * \rm{DTW}(\hat{\boldsymbol{x}}, \boldsymbol{x}) ),
        \end{aligned}
    \label{eq_joint_reward}
\end{equation}
where $\omega$, and $\beta$ are hyper-parameters that control the scale of the reward. We can approximate the expected reward of the predicted path using $N$ sampled paths from the policy $\pi_{\theta_a}$. The derivative of loss function ${\mathcal L}_{\theta} $ is formulated as follows:
\begin{multline}
{\nabla_{{\pi_{\theta}}}\mathcal L_{{\pi_{\theta}}}} = - {\frac{1}{N}} 
\mathop\sum\limits_{i=1}^{N} 
\big[\nabla_{{\pi_{\theta}}} \mathrm{log} \pi_{\theta} 
(a_{i,1},\cdots,a_{i,T}) \times \\
r(a_{i,1},\cdots,a_{i,T})\big]
\label{eq_derivative_reinforcement}
\end{multline}
One limitation of this method is its high variance because it relies solely on samples when calculating the loss function during training. To address this, we use Self Critical Sequence Training (SCST)~\cite{rennie2017self}. In SCST, the algorithm uses a baseline reward $r(\overline{a}_{i,1},\cdots,\overline{a}_{i,T})$ for the $i$-th sample, where $\overline{a}_{i,t}$ is the greedy selection of the model's prediction with frozen parameters. Thus, SCST compares samples from the model against its prediction with fixed parameters. This allows SCST to compare samples against predictions, reducing variance without changing the gradient's expectation~\cite{rennie2017self}. We baseline the gradient using the reward from the arg max of the distribution when generating the next node at each step:

\begin{equation}
    \begin{aligned}
        &\nabla_{{\pi_{\theta}}}\mathcal{L}_{{\pi_{\theta}}} = - \frac{1}{N} \sum_{i=1}^{N} \big[ \nabla_{{\pi_{\theta}}} \log \pi_{\theta}(a_{i,1}, \cdots, a_{i,T}) \times \\
        &\quad \left( r(a_{i,1}, \cdots, a_{i,T}) - r(\overline{a}_{i,1}, \cdots, \overline{a}_{i,T}) \right) \big]
    \end{aligned}
    \label{eq_derivative_scst}
\end{equation}
where $\overline{a}_{i,t}$ is the arg max value of the output distribution for the next node from the $t$-th step of the $i$-th sample. If a sampled sequence is better than the model's greedy selection output, the probability of observing that sample is increased; if the sample is worse, the probability is decreased.

\par Inspired by \cite{hughes2019generating,paulus2017deep}, we also use Maximum-Likelihood Estimation (MLE) objective in path prediction to assist the policy learning algorithm to achieve better prediction. Formally, the joint loss function is formulated as follows:
\begin{equation}
{\mathcal L}_{\theta} = \gamma{\mathcal L_{\pi_{\theta}}} + {\mathcal L_{\rm{CE}}} ,
\label{eq:joint_loss}
\end{equation}
where $\gamma$ is a hyper-parameter. ${\mathcal L_{\rm{CE}}}$ is defined by the MLE objective in path prediction~\cite{GDPICLR24}.

\subsection{MoE guided uncertainty quantification}

\par We propose a MoEUQ (Mixture of Experts for Uncertainty Quantification) model to capture the varying degrees of travel time uncertainty in each segment under different conditions. This segment-level uncertainty is then aggregated to estimate confidence intervals of overall travel time.
\par First, we take the discretized travel time distribution of segments in the predicted path as input, such as \( \boldsymbol{u}^{c_o}_{e_j} = (d_1, f_1, \cdots, d_m, f_m)\) for segment \( e_j\). We select the top-$m$ largest possible travel times in a period before departure time $c_o$. Here, $d_m$ is the $m$-th largest travel time and $f_m$ is the associated probability. We concatenate $\boldsymbol{u}^{c_o}_{e_j}$ with the embeddings of road segments initialized by Node2Vec~\cite{grover2016node2vec} and the time slice of the departure time, then encode them to obtain the embedding $\boldsymbol{r}_j$ of the $j$-th segment. Next, we utilize a wide-deep-recurrent~\cite{ wang2018learning, mao2021estimated} method to model the spatio-temporal dependencies among segments and the influence of other factors. The $j$-th encoded segment in the predicted path is denoted as \(\boldsymbol{r}^{'}_j\), which contains travel time information and aggregated spatio-temporal information of other segments in a path.

\par To better capture travel time uncertainty in each segment, we introduce the Mixture of Experts (MoE) approach~\cite{shazeer2017outrageously}. This method utilizes different combinations of experts to represent each segment based on its aggregated spatio-temporal information, effectively handling varying degrees of travel time uncertainty for the segment under different conditions. Specifically, given the $j$-th encoded segment \(\boldsymbol{r}^{'}_j\), the output of the MoE layer is:

\begin{equation}
\boldsymbol{r}^{''}_j  = \begin{matrix}\sum_{i=1}^{C}G(\boldsymbol{r}^{'}_j)E_{i}(\boldsymbol{r}^{'}_j)\end{matrix},
\label{MoE}
\end{equation}
where \(E_{i}({\boldsymbol{r}^{'}_j})\) denotes the output of the $i$-th expert network, and \(G(\boldsymbol{r}^{'}_j)\) denotes the gating network given \(\boldsymbol{r}^{'}_j\). There are a total $C$ expert networks with separate parameters, and the output of the gating network is a $C$-dimensional vector.
\par To prevent segments from being represented by the same set of experts and to explore better combinations of experts under varying conditions, we implement noisy top-k gating before applying the softmax function in the gating network. Each segment is routed to the most suitable experts, guided by the gating network, to distinguish travel time uncertainty in different conditions. The noisy top-k gating for the $j$-th segment in a predicted path is formulated as:
\begin{equation}
G(\boldsymbol{r}^{'}_j) = \text{Softmax}(\text{TopK}(H(\boldsymbol{r}^{'}_j), k))
\label{Gating}
\end{equation}

\begin{equation}
H(\boldsymbol{r}^{'}_j)_{i} = (\boldsymbol{r}^{'}_j \cdot \boldsymbol{W}_g)_{i} + \mathcal{N}(0, 1) \cdot \text{Softplus}((\boldsymbol{r}^{'}_j \cdot \boldsymbol{W}_{\text{noise}})_i)
\end{equation}

\begin{equation}
    \text{TopK}(\boldsymbol{z}, k)_i =
    \begin{cases} 
        z_i & \text{if } z_i \text{ is in the top } k \text{ elements of } \boldsymbol{z} \\
        -\infty & \text{otherwise}
    \end{cases}
\end{equation}

Next, to estimate the confidence intervals of travel time in the path, we sum along the sequence length dimension based on the output of the MoE layer. Then, we use separate prediction heads to estimate the lower and upper bounds of travel time, supervised by the Mean Interval Score (MIS)~\cite{gneiting2007strictly}. For a confidence level of $1 - \rho$, the predicted upper and lower bounds of the travel time for the $i$-th sample given an ODT query $\boldsymbol{q}$ are defined by $\hat{u}_i = \hat{y}_i + {{\hat{\sigma}^{u}}}_{i}$, and $\hat{l}_i = \hat{y}_i - {{\hat{\sigma}^{l}}}_{i}$, where $\hat{u}_i$ and $\hat{l}_i$ are the $(1 - \frac{\rho}{2})$ and $\frac{\rho}{2}$ quantiles for the $1 - \rho$ confidence interval and $\hat{y}_i$ is the point estimation of travel time, respectively. The MIS loss is formulated as follows:
\begin{equation}
\begin{aligned}
{\mathcal L_{\rm{MIS}}} = \frac{1}{N} & \sum_{i=1}^N \Bigg[ (\hat{u}_i - \hat{l}_i) 
+ \frac{2}{\rho}(y_i - \hat{u}_i) \mathbf{1}\{y_i > \hat{u}_i\} \\
& + \frac{2}{\rho}(\hat{l}_i - y_i) \mathbf{1}{\{y_i < \hat{l}_i\}} 
+ | {y}_i - \hat{y}_i | \Bigg]
\label{eq:MIS_loss}
\end{aligned}
\end{equation}
The MIS loss rewards narrower confidence intervals while encouraging them to include the actual arrival time.

\section{Experiments}

\begin{table*}[t]

    \centering
    \scalebox{0.75}{
        \begin{tabular}{c|ccccc|ccccc}
            \toprule
            Datasets & \multicolumn{5}{c|}{Chengdu} & \multicolumn{5}{c}{Xi'an} \\
            \midrule
            Metric  & RMSE (\(\downarrow\)) & MAE (s) (\(\downarrow\)) & MAPE (\%) (\(\downarrow\)) &  PICP (\%) (\(\uparrow\)) & IW (s) & RMSE (\(\downarrow\)) & MAE (s) (\(\downarrow\)) & MAPE (\%) (\(\downarrow\)) & PICP (\%) (\(\uparrow\)) & IW (s) \\
            \midrule
                     {DOT-dropout}  &   {313.81} & {242.56} & {29.10} & {17.21} & {145.06} & {306.13} & {228.93} & {28.19} & {19.43} & {176.64} \\
     
            {MWSL-TTE}
                &   {307.18} & {239.62} & {29.86} & {73.82} & {782.07}   &   {304.89} & {227.76} & {27.58} & {75.03} & {802.74} \\
          {DeepOD-MIS}
               &   {320.51} & {253.73} & {31.87} & {70.49} & {837.68} &   {319.74} & {251.06} & {30.62}  & {68.26} & {856.26} \\

            {DOT-MIS} &   {315.42} & {242.79} & {30.60} & {71.84} & {806.15}  & {307.15} & {233.26} & {28.70} & {73.25} & {819.51} \\
                      {T-WDR-MIS}
               &   \underline{306.89} & \underline{221.52} & \underline{27.92}& \underline{84.57} & {829.39}  &   \underline{299.13} & \underline{215.61} & \underline{26.81} & \underline{84.93} & {840.28}\\
            \textbf{DutyTTE (ours)} 
                & \textbf{278.22} & \textbf{195.70} & \textbf{23.55} & \textbf{91.02} & {763.65}  & \textbf{270.04}  & \textbf{186.37}  & \textbf{22.95} & \textbf{91.67} & {774.83} \\
              Improvement
               &   {9.34\%} & {11.65\%} & {15.65\%}& {7.63\%} & {-}  &   {9.72\%} & {13.56\%} & {14.39\%} & {7.93\%} & {-}\\
            \bottomrule
        \end{tabular}
    }
    \caption{Performance of travel time uncertainty quantification. }
    \label{tab:uq-results}
        \vspace{-8pt}
\end{table*}

\subsection{Datasets}
We utilize two taxi trajectory datasets collected from from Didi Chuxing\footnote{https://www.didiglobal.com/}. The statistics of datasets are listed in Table~\ref{tab:dataset-statistic}.

\begin{table}[t]
    \setlength{\belowcaptionskip}{-5pt}
    \setlength{\abovecaptionskip}{4pt}
    \centering
    \scalebox{0.80}{
    \begin{tabular}{c|ccc}
      \toprule
      Dataset & Chengdu & Xi'an \\
      \cmidrule(lr){1-3} 
      Time span & 11.01--11.16, 2018 & 11.01--11.16, 2018 \\
      Sampling rate & 3s & 3s \\
      Number of trajectories & 1,613,355 & 1,951,585 \\
          average travel time & 810.18s & 770.32s \\
      Area (width$\ast$height km$^2$) & 
        14.38$\ast$9.98 & 10.14$\ast$8.87 \\
      Number of nodes  & 
        4599 & 5259 \\
        Number of edges  & 
        6737 & 7784 \\
      \bottomrule
    \end{tabular}
    }
    \caption{Dataset statistics.}
    \label{tab:dataset-statistic}
        \vspace{-4pt}
\end{table}

\subsection{Comparison Methods}
First, baselines for travel time uncertainty quantification include: \textbf{(1) T-WDR-MIS}: Using a transformer for path prediction, with WDR~\cite{WDR} for TTE. It uses prediction heads to output confidence intervals, trained with MIS loss. \textbf{(2) MWSL-TTE}~\cite{wang2023multi}: Sums estimated travel time across all segments of the predicted path. Uncertainty is based on variance from a learned Gaussian distribution for each segment. \textbf{(3) DeepOD-MIS}: Uses DeepOD~\cite{yuan2020effective} with prediction heads for confidence intervals, trained with MIS loss. \textbf{(4) DOT-dropout}: Incorporates MC Dropout~\cite{zhu2017deep} into DOT~\cite{DOT-SIGMOD23}. \textbf{(5) DOT-MIS}: Equips DOT with prediction heads for confidence intervals, trained with MIS.

\par The desired confidence level of the confidence interval for travel time is set at $90 \%$~\cite{angelopoulos2022image, bates2021distribution}. To evaluate uncertainty quantification, we use Interval Width (IW) and Prediction Interval Coverage Probability (PICP)~\cite{lawless2005frequentist, zhou2021stuanet}, where PICP is defined as the proportion of ground truth values covered by the estimated intervals.

\par Additional baselines for TTE: \textbf{(1) TEMP}~\cite{wang2019simple}: Estimate travel time using neighboring trips of a query. \textbf{(2) GBDT}~\cite{chen2016xgboost}: A machine learning method. \textbf{(3) ST-NN}:~\cite{jindal2017unified}: Jointly predicts travel distance and time. 

\par We also compare methods for path prediction: \textbf{(1) Dijkstra's algorithm (DA)}:~\cite{johnson1973note} Searches for the shortest path for given OD pairs. \textbf{(2) NMLR}~\cite{jain2021neuromlr}: Learns segment weights and searches for paths with the largest weight at each step. \textbf{(3) Key Segment (KS)}~\cite{tian2023effective}: Detects a relay vertex for given OD pairs, predicting paths from origin to relay, then to the destination. \textbf{(4) Transformer}: Utilizes a transformer~\cite{vaswani2017attention} for path prediction as implemented in~\cite{GDPICLR24}. \textbf{(5) GDP}~\cite{GDPICLR24}: uses a diffusion-based model to learn path distributions and predict paths via transformers.

\begin{table}[t]
    \setlength{\belowcaptionskip}{-5pt}
    \setlength{\abovecaptionskip}{4pt}
    \centering
    \scriptsize 
    \begin{tabular}{c|ccc}
        \toprule
        Datasets & \multicolumn{3}{c}{Chengdu~/~Xi'an} \\
        \midrule
        Metric & RMSE (\(\downarrow\)) & MAE (s) (\(\downarrow\)) & MAPE (\%) (\(\downarrow\)) \\
        \midrule
        {TEMP}
            & {317.33/312.06} & {259.17/249.62} & {32.16/32.07} \\
        {GBDT}
            & {315.91/318.10} & {247.03/245.83} & {31.03/30.12} \\
        {STNN}
            & {324.62/336.56} & {268.67/264.95} & {33.28/32.93} \\
        {DeepOD}
            & {320.51/319.74} & {253.73/251.06} & {31.87/30.62} \\
        {MWSL-TTE}
            & {307.18/304.89} & {239.62/227.76} & {29.86/27.58} \\
        {DOT}
            & {315.42/307.15} & {242.79/233.26} & {30.60/28.70} \\
        {T-WDR}
            & {\underline{306.89}/\underline{299.13}} & {\underline{221.52}/\underline{215.61}} & {\underline{27.92}/\underline{26.81}} \\
        \textbf{DutyTTE (ours)} & \textbf{278.22/270.04}  & \textbf{195.70/186.37}  & \textbf{23.55/22.95} \\
        Improvement & 9.34\%/9.72\%  & 11.65\%/13.56\%  & 15.65\%/14.39\% \\
        \bottomrule
    \end{tabular}
    \caption{Performance of travel time estimation. }
     \vspace{-13pt}
    \label{tab:travel-time-estimation-result}
\end{table}

\begin{table}[t]
    \centering
    \scalebox{0.88}{
        \begin{tabular}{c|cc|cc}
            \toprule
            Datasets & \multicolumn{2}{c|}{Chengdu} & \multicolumn{2}{c}{Xi'an} \\
            \midrule
            Metric & LCS (\(\uparrow\)) & DTW (\(\downarrow\)) & LCS (\(\uparrow\)) & DTW (\(\downarrow\)) \\
            \midrule
            {DA}
                & {3.74} & {0.103} & {4.62} & {0.093} \\
            {NMLR}
                & {6.09} & {0.112} & {7.19} & {0.102} \\
            {KS}
                & {12.03} & {0.039} & {12.82} & {0.036} \\
           {Transformer}
                & {13.48} & {0.019} & {13.85} & {0.024} \\
            {GDP}
                & \underline{13.51} & \underline{0.019} & \underline{13.91} & \underline{0.024} \\
  
            \textbf{DutyTTE}
                & \textbf{14.52} & \textbf{0.017} & \textbf{14.64} & \textbf{0.021} \\
             Improvement & 7.48\% & 10.53\%  & 5.24\% & 12.50\%  \\
            \bottomrule
        \end{tabular}
    }
    \caption{Performance of path prediction. }
    \label{tab:path-prediction-result}
    \vspace{-8pt}
\end{table}

\subsection{Overall Comparison}

\subsubsection{Effectiveness of travel time uncertainty quantification} 

T-WDR-MIS and MWSL-TTE outperform DeepOD-MIS in PICP by leveraging estimated travel time in the predicted path, highlighting the importance of predicting paths before quantifying uncertainty. T-WDR-MIS predicts paths autoregressively by maximizing the likelihood of each segment independently, while MWSL-TTE estimates transition weights between segments to predict paths. Both methods suffer from error accumulation in path prediction, limiting their ability to provide reliable path information for uncertainty quantification.

In contrast, DutyTTE enhances alignment with ground truth using DRL techniques to predict accurate paths while avoiding untraveled segment data. It employs MoEUQ to capture segment-level uncertainty in complex contexts by adaptively selecting experts to handle varying conditions. This enables DutyTTE to effectively learn the contribution of each segment's travel time to overall uncertainty, achieving superior performance.

\subsubsection{Effectiveness of travel time estimation}
Table~\ref{tab:travel-time-estimation-result} presents the TTE results, where DutyTTE consistently outperforms other baselines across two datasets.

MWSL-TTE and T-WDR are more accurate than DOT and DeepOD by predicting potential paths and utilizing recent travel time information. T-WDR outperforms MWSL-TTE; while MWSL-TTE uses weak supervision for segment travel times, T-WDR leverages recent data in road segments and its wide-deep-recurrent architecture to capture correlations among influencing factors.

DutyTTE improves MAPE by 14.39\% to 15.65\% over the most competitive baseline. This is due to precise path prediction between OD pairs and a mixture of experts guided travel time estimation method that effectively utilizes road segment travel time distributions under varying contexts.

\subsubsection{Effectiveness of path prediction}
We compare the accuracy of path prediction using DTW~\cite{muller2007dynamic} and LCS~\cite{bergroth2000survey} to measure the similarity between predicted paths and the ground truth. A smaller DTW or a larger LCS indicates better performance. The results are shown in Table~\ref{tab:path-prediction-result}.

\begin{figure}[!t]
    \centering
    \includegraphics[width= .75\linewidth]{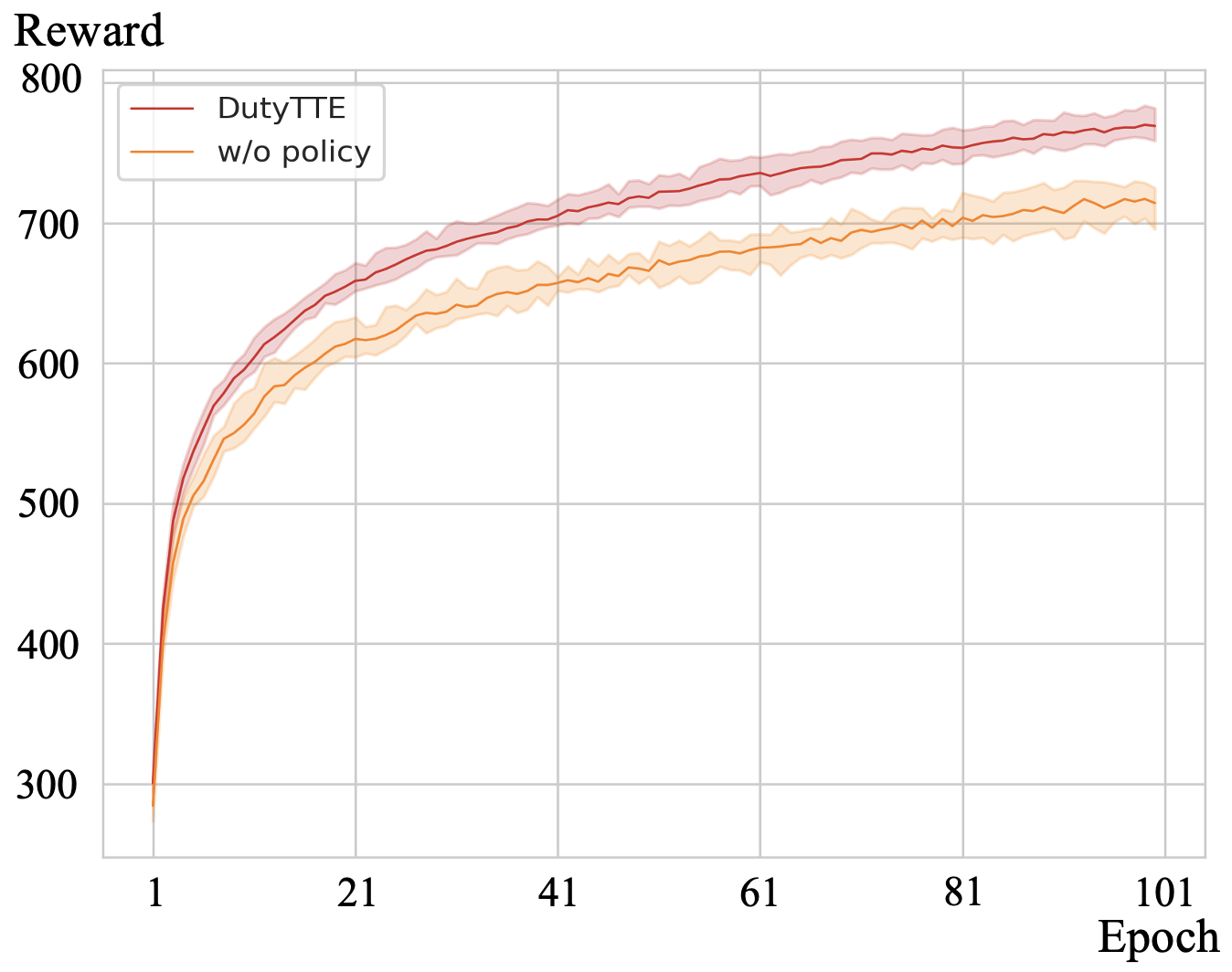}
    \caption{{Reward curves during training.}}
    \label{fig:reward}
    \vspace{-12pt}
\end{figure}

{
\par KS surpasses DA and NMLR by identifying key segments and predicting paths based on constructed OD pairs with shorter distances. Both KS and NMLR rely on search-based methods that generate the next segment based solely on the learned weights among road segments. GDP models transition patterns in road networks using a diffusion model and uses a transformer to predict paths, yet still struggles to address error accumulation. DutyTTE addresses this by using a DRL method to enhance alignment between predicted and true paths, achieving superior performance. Figure~\ref{fig:reward} illustrates the reward curves for DutyTTE and a transformer implemented in GDP~\cite{GDPICLR24} trained without policy loss (w/o policy). The higher rewards achieved by DutyTTE demonstrate enhanced path alignment resulting from DRL techniques.
}

\begin{table}[t]
    \centering
    \scalebox{0.75}{
        \begin{tabular}{c|ccccc}
            \toprule
            Dataset & \multicolumn{5}{c}{Chengdu}\\
            \midrule
            Metric  & RMSE (\(\downarrow\)) & MAE(\(\downarrow\)) & MAPE (\(\downarrow\)) & PICP (\(\uparrow\)) & IW   \\
            \midrule
            {w/o-M} 
                & {282.41} & {198.47} & {24.32} &  {88.63} & {781.52} \\
            {w/o-P} 
                & {302.93} & {219.32} & {27.28} & {87.15} &  {836.09}  \\
            \midrule
            {DutyTTE}
                & {278.22} & {195.70} & {23.55} & {91.02} & {763.65}  \\
            \bottomrule
        \end{tabular}      
    }  
    \caption{Ablation study.}
    \label{tab:ablation-study}
      \vspace{-6pt}
\end{table}

\subsubsection{Ablation Study}
\par To verify the effectiveness of components, we conduct an ablation study with the following DutyTTE variants: 1) w/o-M: removing the mixture of experts guided uncertainty quantification mechanism. 2) w/o-P: removing the policy loss and using only cross-entropy loss in path prediction.
\par We compare the performance of these variants with DutyTTE, and the experimental results are presented in Table~\ref{tab:ablation-study}. We observe the following: 1) The MoEUQ is crucial for accurate uncertainty quantification, as MoEUQ effectively learns the uncertainty of each segment under different contexts. 2) Integrating metrics that measure the overall similarity between predicted paths and the ground truth into the policy loss during training leads to improvements in path prediction performance.

\subsubsection{Hyper-parameter Analysis}

\begin{figure}[!t]
    \centering
    \includegraphics[width= 1\linewidth]{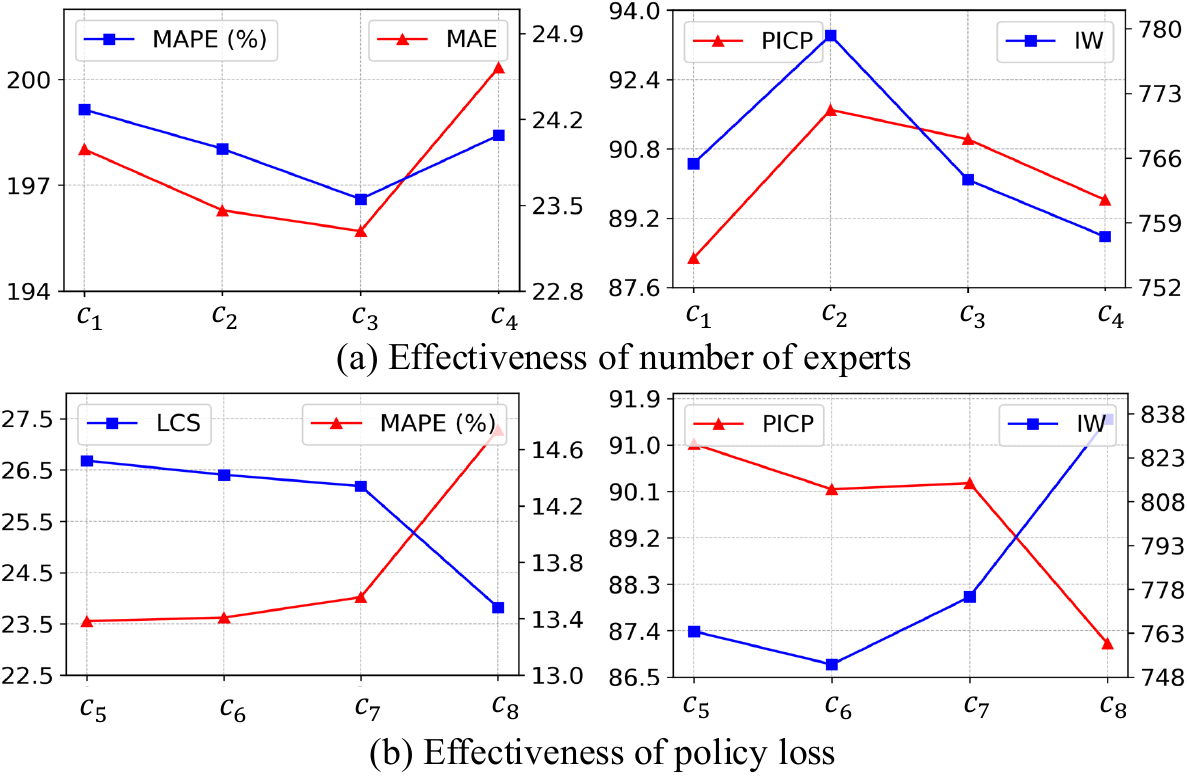}
    \caption{{Effectiveness of hyper-parameters.}}
    \label{fig:hyper-params}
    \vspace{-8pt}

\end{figure}

 We demonstrate the effectiveness of key hyper-parameters on testing sets. In Figure~\ref{fig:hyper-params} (a), we set the value of $k$ and the number of experts as follows: \(c_1: k=1, C=8\), \(c_2: k=2, C=8\), \(c_3: k=4, C=8\), \(c_4: k=6, C=8\). In Figure~\ref{fig:hyper-params} (b), we configure the policy loss function with the following hyper-parameters: \(c_5: \omega = 1, \beta = 50, \gamma = 1\), \(c_6: \omega = 1, \beta = 100, \gamma = 1\), \(c_7: \omega = 1, \beta = 50, \gamma = 0.5\), \(c_8: \omega = 0, \beta = 0, \gamma = 0\). We observe the following results.

\par 1) In Figure~\ref{fig:hyper-params} (a), setting the value of $k$ to 4 and the number of experts to $8$ can lead to optimal performance. This can be attributed to the fact that having too few $k$ and experts limits MoEUQ's capability, as it can only handle a limited number of contexts for evaluating travel time uncertainty in a segment. However, if the number of experts is too large, the model complexity also increases, complicating the training process and  compromising performance.

\par 2) In Figure~\ref{fig:hyper-params} (b), applying the policy loss function with different parameter settings, such as \(c_5\), \(c_6\), and \(c_7\), results in improved performance compared to not using the policy loss function, as seen with \(c_8\). Among these configurations, \(c_5\) stands out as a better option for more accurate path prediction compared to \(c_6\) and \(c_7\). Achieving more accurate path predictions also leads to improvements in travel-time estimation and uncertainty quantification.

\subsubsection{Analysis of the MoE guided uncertainty quantification}
\begin{figure}[!t]
    \centering
    \includegraphics[width= 1\linewidth]{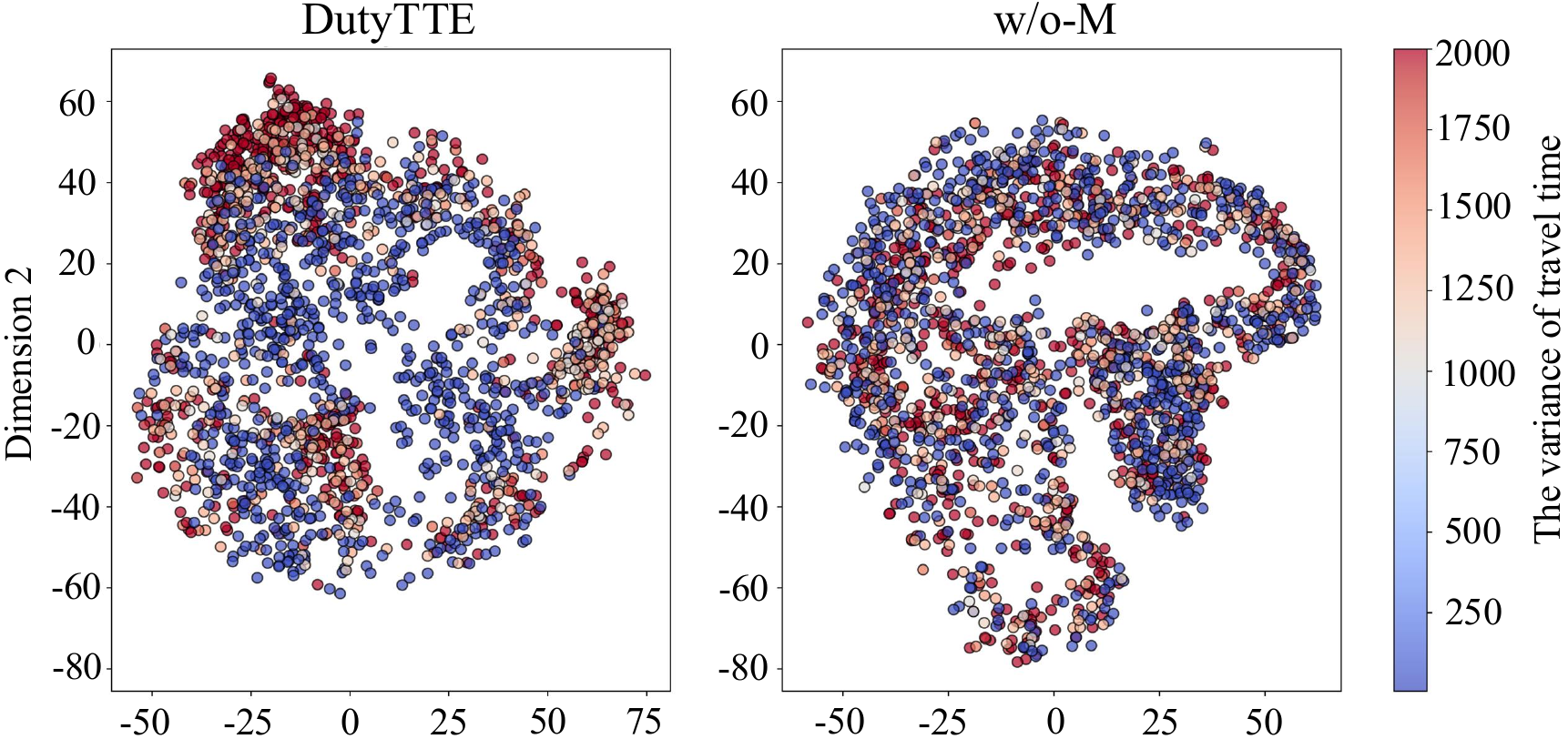}
    \caption{{t-SNE Visualization for embeddings of road segments under varying contexts.}}
    \label{fig:emb-vis}
\end{figure}

\begin{figure}[!t]
    \centering
    \includegraphics[width= 1\linewidth]{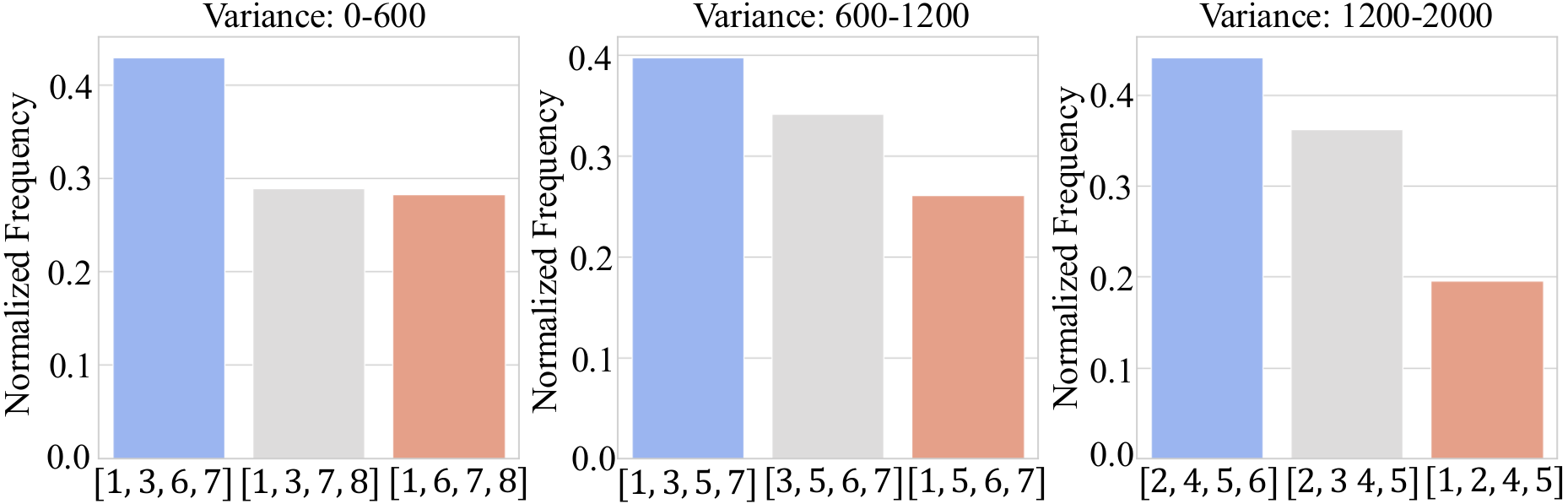}
    \caption{{Expert activation distributions.}}
    \label{fig:expert-activation}
        \vspace{-8pt}
\end{figure}

The experimental results in Figures~\ref{fig:emb-vis} and Figure~\ref{fig:expert-activation} demonstrate the effectiveness of the MoE method in quantifying uncertainty for road segment travel times. In Figure~\ref{fig:emb-vis}, 2000 road segment samples under various traffic conditions were randomly selected, with travel time variance calculated from historical trajectories observed 10 minutes before trip departure. This variance reflects the uncertainty of travel time for each segment. The t-SNE visualization reduces high-dimensional embeddings of these segments into two dimensions for comparison. The results demonstrate that with MoE, clusters of road segments with low and high travel time variance (indicating lower and higher uncertainty) are more distinct compared to the scenario without MoE, where cluster boundaries are less defined. This underscores MoE's ability to better capture and differentiate travel time uncertainty under varying traffic conditions, improving overall uncertainty quantification. Additionally, Figure~\ref{fig:expert-activation} shows the most frequently activated combination of experts for road segments with varying levels of uncertainty. The distribution of activated experts differs across uncertainty levels, indicating that the model dynamically employs diverse combinations of experts to effectively capture the uncertainty characteristics of road segments under different traffic conditions.

\begin{table}[t]
    \centering
    \scalebox{0.77}{
        \begin{tabular}{c|cccccc}
            \toprule
            Dataset & DA & DOT & T-WDR & MWSL & GDP & DutyTTE \\
            \midrule
            Chengdu & 0.0722 & 1.362 & 0.2675 & 0.1289 & 5.093 & 0.2706 \\
            Xi'an & 0.0718 & 1.293 & 0.2649 & 0.1357 & 4.896 & 0.2682 \\
            \bottomrule
        \end{tabular}
    }

    \caption{Execution time comparison (seconds per batch).}
        \label{tab:efficiency_study}
    \vspace{-8pt}
\end{table}

\subsubsection{Efficiency Study}
We evaluate the average execution time per batch, consisting of 128 samples from Chengdu and Xi'an, for efficiency validation, as shown in Table~\ref{tab:efficiency_study}. DutyTTE uses a DRL method for training; during prediction, the policy network allows for fast inferences. Given DutyTTE's good performance, we see this as a promising trade-off between efficiency and effectiveness.

\section{Conclusion}
We introduce DutyTTE for quantifying uncertainty in OD travel times. Using a DRL approach, we optimize path prediction to align closely with the ground truth and enhance uncertainty quantification. To capture travel time uncertainty across road segments, we propose a mixture of experts mechanism that models the impact of segments in complex contexts. Experiments on two datasets validate DutyTTE's effectiveness in path prediction, travel time uncertainty quantification.

\section{Acknowledgment}
This work was supported by the Fundamental Research Funds for the Central Universities (Grant No. 2023YJS030).

\bibliography{aaai25}

\begin{thebibliography}{50}
\providecommand{\natexlab}[1]{#1}

\bibitem[{Angelopoulos et~al.(2022)Angelopoulos, Kohli, Bates, Jordan, Malik, Alshaabi, Upadhyayula, and Romano}]{angelopoulos2022image}
Angelopoulos, A.~N.; Kohli, A.~P.; Bates, S.; Jordan, M.; Malik, J.; Alshaabi, T.; Upadhyayula, S.; and Romano, Y. 2022.
\newblock Image-to-image regression with distribution-free uncertainty quantification and applications in imaging.
\newblock In \emph{International Conference on Machine Learning}, 717--730. PMLR.

\bibitem[{Bates et~al.(2021)Bates, Angelopoulos, Lei, Malik, and Jordan}]{bates2021distribution}
Bates, S.; Angelopoulos, A.; Lei, L.; Malik, J.; and Jordan, M. 2021.
\newblock Distribution-free, risk-controlling prediction sets.
\newblock \emph{Journal of the ACM (JACM)}, 68(6): 1--34.

\bibitem[{Bergroth, Hakonen, and Raita(2000)}]{bergroth2000survey}
Bergroth, L.; Hakonen, H.; and Raita, T. 2000.
\newblock A survey of longest common subsequence algorithms.
\newblock In \emph{Proceedings Seventh International Symposium on String Processing and Information Retrieval. SPIRE 2000}, 39--48. IEEE.

\bibitem[{Chen and Guestrin(2016)}]{chen2016xgboost}
Chen, T.; and Guestrin, C. 2016.
\newblock Xgboost: A scalable tree boosting system.
\newblock In \emph{ACM SIGKDD}, 785--794.

\bibitem[{Chen et~al.(2022)Chen, Xiao, Gong, Fang, Ma, Chai, and Cao}]{chen2022interpreting}
Chen, Z.; Xiao, X.; Gong, Y.-J.; Fang, J.; Ma, N.; Chai, H.; and Cao, Z. 2022.
\newblock Interpreting trajectories from multiple views: A hierarchical self-attention network for estimating the time of arrival.
\newblock In \emph{Proceedings of the 28th ACM SIGKDD Conference on Knowledge Discovery and Data Mining}, 2771--2779.

\bibitem[{Derrow-Pinion et~al.(2021)Derrow-Pinion, She, Wong, Lange, Hester, Perez, Nunkesser, Lee, Guo, Wiltshire et~al.}]{derrow2021eta}
Derrow-Pinion, A.; She, J.; Wong, D.; Lange, O.; Hester, T.; Perez, L.; Nunkesser, M.; Lee, S.; Guo, X.; Wiltshire, B.; et~al. 2021.
\newblock Eta prediction with graph neural networks in google maps.
\newblock In \emph{Proceedings of the 30th ACM International Conference on Information \& Knowledge Management}, 3767--3776.

\bibitem[{Fang et~al.(2020)Fang, Huang, Wang, Zeng, Liang, and Wang}]{fang2020constgat}
Fang, X.; Huang, J.; Wang, F.; Zeng, L.; Liang, H.; and Wang, H. 2020.
\newblock Constgat: Contextual spatial-temporal graph attention network for travel time estimation at baidu maps.
\newblock In \emph{Proceedings of the 26th ACM SIGKDD International Conference on Knowledge Discovery \& Data Mining}, 2697--2705.

\bibitem[{Fu et~al.(2020)Fu, Meng, Ye, and Wang}]{CompactETA}
Fu, K.; Meng, F.; Ye, J.; and Wang, Z. 2020.
\newblock CompactETA: A fast inference system for travel time prediction.
\newblock In \emph{Proceedings of the 26th ACM SIGKDD International Conference on Knowledge Discovery \& Data Mining}, 3337--3345.

\bibitem[{Gasthaus et~al.(2019)Gasthaus, Benidis, Wang, Rangapuram, Salinas, Flunkert, and Januschowski}]{gasthaus2019probabilistic}
Gasthaus, J.; Benidis, K.; Wang, Y.; Rangapuram, S.~S.; Salinas, D.; Flunkert, V.; and Januschowski, T. 2019.
\newblock Probabilistic forecasting with spline quantile function RNNs.
\newblock In \emph{The 22nd international conference on artificial intelligence and statistics}, 1901--1910. PMLR.

\bibitem[{Gneiting and Raftery(2007)}]{gneiting2007strictly}
Gneiting, T.; and Raftery, A.~E. 2007.
\newblock Strictly proper scoring rules, prediction, and estimation.
\newblock \emph{Journal of the American statistical Association}, 102(477): 359--378.

\bibitem[{Grover and Leskovec(2016)}]{grover2016node2vec}
Grover, A.; and Leskovec, J. 2016.
\newblock node2vec: Scalable feature learning for networks.
\newblock In \emph{Proceedings of the 22nd ACM SIGKDD international conference on Knowledge discovery and data mining}, 855--864.

\bibitem[{Hochreiter and Schmidhuber(1997)}]{hochreiter1997long}
Hochreiter, S.; and Schmidhuber, J. 1997.
\newblock Long short-term memory.
\newblock \emph{Neural computation}, 9(8): 1735--1780.

\bibitem[{Hoeffding(1994)}]{hoeffding1994probability}
Hoeffding, W. 1994.
\newblock Probability inequalities for sums of bounded random variables.
\newblock \emph{The collected works of Wassily Hoeffding}, 409--426.

\bibitem[{Hong et~al.(2020)Hong, Lin, Yang, Li, Fu, Wang, Qie, and Ye}]{hong2020heteta}
Hong, H.; Lin, Y.; Yang, X.; Li, Z.; Fu, K.; Wang, Z.; Qie, X.; and Ye, J. 2020.
\newblock HetETA: Heterogeneous information network embedding for estimating time of arrival.
\newblock In \emph{Proceedings of the 26th ACM SIGKDD international conference on knowledge discovery \& data mining}, 2444--2454.

\bibitem[{Hughes, Chang, and Zhang(2019)}]{hughes2019generating}
Hughes, J.~W.; Chang, K.-h.; and Zhang, R. 2019.
\newblock Generating better search engine text advertisements with deep reinforcement learning.
\newblock In \emph{Proceedings of the 25th ACM SIGKDD International Conference on Knowledge Discovery \& Data Mining}, 2269--2277.

\bibitem[{Izmailov et~al.(2021)Izmailov, Vikram, Hoffman, and Wilson}]{pmlr-v139-izmailov21a}
Izmailov, P.; Vikram, S.; Hoffman, M.~D.; and Wilson, A. G.~G. 2021.
\newblock What Are Bayesian Neural Network Posteriors Really Like?
\newblock In Meila, M.; and Zhang, T., eds., \emph{Proceedings of the 38th International Conference on Machine Learning}, volume 139 of \emph{Proceedings of Machine Learning Research}, 4629--4640. PMLR.

\bibitem[{Jain et~al.(2021)Jain, Bagadia, Manchanda, and Ranu}]{jain2021neuromlr}
Jain, J.; Bagadia, V.; Manchanda, S.; and Ranu, S. 2021.
\newblock Neuromlr: Robust \& reliable route recommendation on road networks.
\newblock \emph{Advances in Neural Information Processing Systems}, 34: 22070--22082.

\bibitem[{Jin et~al.(2022)Jin, Wang, Zhang, Sha, and Huang}]{jin2022stgnn}
Jin, G.; Wang, M.; Zhang, J.; Sha, H.; and Huang, J. 2022.
\newblock STGNN-TTE: Travel time estimation via spatial--temporal graph neural network.
\newblock \emph{Future Generation Computer Systems}, 126: 70--81.

\bibitem[{Jindal et~al.(2017)Jindal, Chen, Nokleby, Ye et~al.}]{jindal2017unified}
Jindal, I.; Chen, X.; Nokleby, M.; Ye, J.; et~al. 2017.
\newblock A unified neural network approach for estimating travel time and distance for a taxi trip.
\newblock \emph{arXiv preprint arXiv:1710.04350}.

\bibitem[{Johnson(1973)}]{johnson1973note}
Johnson, D.~B. 1973.
\newblock A note on Dijkstra's shortest path algorithm.
\newblock \emph{Journal of the ACM}, 20(3): 385--388.

\bibitem[{Kendall and Gal(2017)}]{kendall2017uncertainties}
Kendall, A.; and Gal, Y. 2017.
\newblock What uncertainties do we need in bayesian deep learning for computer vision?
\newblock \emph{Advances in neural information processing systems}, 30.

\bibitem[{Kong, Sun, and Zhang(2020)}]{kong2020sde}
Kong, L.; Sun, J.; and Zhang, C. 2020.
\newblock Sde-net: Equipping deep neural networks with uncertainty estimates.
\newblock \emph{arXiv preprint arXiv:2008.10546}.

\bibitem[{Lakshminarayanan, Pritzel, and Blundell(2017)}]{lakshminarayanan2017simple}
Lakshminarayanan, B.; Pritzel, A.; and Blundell, C. 2017.
\newblock Simple and scalable predictive uncertainty estimation using deep ensembles.
\newblock \emph{Advances in neural information processing systems}, 30.

\bibitem[{Lawless and Fredette(2005)}]{lawless2005frequentist}
Lawless, J.~F.; and Fredette, M. 2005.
\newblock Frequentist prediction intervals and predictive distributions.
\newblock \emph{Biometrika}, 92(3): 529--542.

\bibitem[{Li et~al.(2019)Li, Cong, Sun, and Cheng}]{li2019learning}
Li, X.; Cong, G.; Sun, A.; and Cheng, Y. 2019.
\newblock Learning travel time distributions with deep generative model.
\newblock In \emph{WWW}, 1017--1027.

\bibitem[{Li et~al.(2018)Li, Fu, Wang, Shahabi, Ye, and Liu}]{li2018multi}
Li, Y.; Fu, K.; Wang, Z.; Shahabi, C.; Ye, J.; and Liu, Y. 2018.
\newblock Multi-task representation learning for travel time estimation.
\newblock In \emph{ACM SIGKDD}, 1695--1704.

\bibitem[{Lin et~al.(2023)Lin, Wan, Hu, Guo, Yang, Lin, and Jensen}]{DOT-SIGMOD23}
Lin, Y.; Wan, H.; Hu, J.; Guo, S.; Yang, B.; Lin, Y.; and Jensen, C.~S. 2023.
\newblock Origin-Destination Travel Time Oracle for Map-Based Services.
\newblock \emph{Proc. ACM Manag. Data}, 1(3).

\bibitem[{Liu et~al.(2023)Liu, Jiang, Liu, and Chen}]{liu2023uncertainty}
Liu, H.; Jiang, W.; Liu, S.; and Chen, X. 2023.
\newblock Uncertainty-Aware Probabilistic Travel Time Prediction for On-Demand Ride-Hailing at DiDi.
\newblock In \emph{Proceedings of the 29th ACM SIGKDD Conference on Knowledge Discovery and Data Mining}, 4516--4526.

\bibitem[{Liu et~al.(2020)Liu, Qin, Zhang, Pei, Jiang, Feng, and Zhou}]{liu2020probabilistic}
Liu, Y.; Qin, H.; Zhang, Z.; Pei, S.; Jiang, Z.; Feng, Z.; and Zhou, J. 2020.
\newblock Probabilistic spatiotemporal wind speed forecasting based on a variational Bayesian deep learning model.
\newblock \emph{Applied Energy}, 260: 114259.

\bibitem[{M{\"u}ller(2007)}]{muller2007dynamic}
M{\"u}ller, M. 2007.
\newblock Dynamic time warping.
\newblock \emph{Information retrieval for music and motion}, 69--84.

\bibitem[{Paulus, Xiong, and Socher(2017)}]{paulus2017deep}
Paulus, R.; Xiong, C.; and Socher, R. 2017.
\newblock A deep reinforced model for abstractive summarization.
\newblock \emph{arXiv preprint arXiv:1705.04304}.

\bibitem[{Rennie et~al.(2017)Rennie, Marcheret, Mroueh, Ross, and Goel}]{rennie2017self}
Rennie, S.~J.; Marcheret, E.; Mroueh, Y.; Ross, J.; and Goel, V. 2017.
\newblock Self-critical sequence training for image captioning.
\newblock In \emph{Proceedings of the IEEE conference on computer vision and pattern recognition}, 7008--7024.

\bibitem[{Shazeer et~al.(2017)Shazeer, Mirhoseini, Maziarz, Davis, Le, Hinton, and Dean}]{shazeer2017outrageously}
Shazeer, N.; Mirhoseini, A.; Maziarz, K.; Davis, A.; Le, Q.; Hinton, G.; and Dean, J. 2017.
\newblock Outrageously large neural networks: The sparsely-gated mixture-of-experts layer.
\newblock \emph{arXiv preprint arXiv:1701.06538}.

\bibitem[{Shi et~al.(2024)Shi, Tong, Zhou, Xu, Wang, and Ye}]{GDPICLR24}
Shi, D.; Tong, Y.; Zhou, Z.; Xu, K.; Wang, Z.; and Ye, J. 2024.
\newblock GRAPH-CONSTRAINED DIFFUSION FOR END-TO-END PATH PLANNING.
\newblock In \emph{The Twelfth International Conference on Learning Representations}.

\bibitem[{Tian et~al.(2023)Tian, Shi, Luo, Li, Xie, and Zou}]{tian2023effective}
Tian, W.; Shi, J.; Luo, S.; Li, H.; Xie, X.; and Zou, Y. 2023.
\newblock Effective and efficient route planning using historical trajectories on road networks.
\newblock \emph{Proceedings of the VLDB Endowment}, 16(10): 2512--2524.

\bibitem[{Vaswani et~al.(2017)Vaswani, Shazeer, Parmar, Uszkoreit, Jones, Gomez, Kaiser, and Polosukhin}]{vaswani2017attention}
Vaswani, A.; Shazeer, N.; Parmar, N.; Uszkoreit, J.; Jones, L.; Gomez, A.~N.; Kaiser, {\L}.; and Polosukhin, I. 2017.
\newblock Attention is all you need.
\newblock \emph{Advances in neural information processing systems}, 30.

\bibitem[{Wang et~al.(2018)Wang, Zhang, Cao, Li, and Zheng}]{wang2018will}
Wang, D.; Zhang, J.; Cao, W.; Li, J.; and Zheng, Y. 2018.
\newblock When will you arrive? Estimating travel time based on deep neural networks.
\newblock In \emph{AAAI}, 2500--2507.

\bibitem[{Wang et~al.(2019)Wang, Tang, Kuo, Kifer, and Li}]{wang2019simple}
Wang, H.; Tang, X.; Kuo, Y.-H.; Kifer, D.; and Li, Z. 2019.
\newblock A simple baseline for travel time estimation using large-scale trip data.
\newblock \emph{ACM Trans. on Intelli. Sys. and Tech.}, 10(2): 1--22.

\bibitem[{Wang et~al.(2023)Wang, Zhang, Fan, Chen, Zhang, Shibasaki, and Song}]{wang2023multi}
Wang, H.; Zhang, Z.; Fan, Z.; Chen, J.; Zhang, L.; Shibasaki, R.; and Song, X. 2023.
\newblock Multi-Task Weakly Supervised Learning for Origin--Destination Travel Time Estimation.
\newblock \emph{IEEE Transactions on Knowledge and Data Engineering}, 35(11): 11628--11641.

\bibitem[{Wang, Fu, and Ye(2018{\natexlab{a}})}]{wang2018learning}
Wang, Z.; Fu, K.; and Ye, J. 2018{\natexlab{a}}.
\newblock Learning to estimate the travel time.
\newblock In \emph{ACM SIGKDD}, 858--866.

\bibitem[{Wang, Fu, and Ye(2018{\natexlab{b}})}]{WDR}
Wang, Z.; Fu, K.; and Ye, J. 2018{\natexlab{b}}.
\newblock Learning to estimate the travel time.
\newblock In \emph{Proceedings of the 24th ACM SIGKDD international conference on knowledge discovery \& data mining}, 858--866.

\bibitem[{Wu et~al.(2021)Wu, Gao, Chinazzi, Xiong, Vespignani, Ma, and Yu}]{wu2021quantifying}
Wu, D.; Gao, L.; Chinazzi, M.; Xiong, X.; Vespignani, A.; Ma, Y.-A.; and Yu, R. 2021.
\newblock Quantifying uncertainty in deep spatiotemporal forecasting.
\newblock In \emph{Proceedings of the 27th ACM SIGKDD Conference on Knowledge Discovery \& Data Mining}, 1841--1851.

\bibitem[{Xu, Wang, and Sun(2024)}]{xu2024link}
Xu, C.; Wang, Q.; and Sun, L. 2024.
\newblock Link Representation Learning for Probabilistic Travel Time Estimation.
\newblock \emph{arXiv preprint arXiv:2407.05895}.

\bibitem[{Xu et~al.(2020)Xu, Xu, Zhou, Liu, Li, and Liu}]{xu2020tadnm}
Xu, S.; Xu, J.; Zhou, R.; Liu, C.; Li, Z.; and Liu, A. 2020.
\newblock Tadnm: A transportation-mode aware deep neural model for travel time estimation.
\newblock In \emph{DASFAA}, 468--484.

\bibitem[{Xu et~al.(2018)Xu, Li, Guan, Zhang, Li, Nan, Liu, Bian, and Ye}]{xu2018large}
Xu, Z.; Li, Z.; Guan, Q.; Zhang, D.; Li, Q.; Nan, J.; Liu, C.; Bian, W.; and Ye, J. 2018.
\newblock Large-scale order dispatch in on-demand ride-hailing platforms: A learning and planning approach.
\newblock In \emph{Proceedings of the 24th ACM SIGKDD international conference on knowledge discovery \& data mining}, 905--913.

\bibitem[{Yuan, Li, and Bao(2022)}]{yuan2022route}
Yuan, H.; Li, G.; and Bao, Z. 2022.
\newblock Route travel time estimation on a road network revisited: Heterogeneity, proximity, periodicity and dynamicity.
\newblock \emph{Proceedings of the VLDB Endowment}, 16(3): 393--405.

\bibitem[{Yuan et~al.(2020{\natexlab{a}})Yuan, Li, Bao, and Feng}]{DeepOD}
Yuan, H.; Li, G.; Bao, Z.; and Feng, L. 2020{\natexlab{a}}.
\newblock Effective Travel Time Estimation: When Historical Trajectories over Road Networks Matter.
\newblock In \emph{SIGMOD}, 2135--2149.

\bibitem[{Yuan et~al.(2020{\natexlab{b}})Yuan, Li, Bao, and Feng}]{yuan2020effective}
Yuan, H.; Li, G.; Bao, Z.; and Feng, L. 2020{\natexlab{b}}.
\newblock Effective travel time estimation: When historical trajectories over road networks matter.
\newblock In \emph{Proceedings of the 2020 acm sigmod international conference on management of data}, 2135--2149.

\bibitem[{Zhou et~al.(2021)Zhou, Wang, Xie, Qiao, and Li}]{zhou2021stuanet}
Zhou, Z.; Wang, Y.; Xie, X.; Qiao, L.; and Li, Y. 2021.
\newblock Stuanet: Understanding uncertainty in spatiotemporal collective human mobility.
\newblock In \emph{Proceedings of the Web Conference 2021}, 1868--1879.

\bibitem[{Zhu and Laptev(2017)}]{zhu2017deep}
Zhu, L.; and Laptev, N. 2017.
\newblock Deep and confident prediction for time series at uber.
\newblock In \emph{2017 IEEE International Conference on Data Mining Workshops (ICDMW)}, 103--110. IEEE.

\end{thebibliography}

\end{document}